\pdfoutput=1
\documentclass[10pt,twocolumn,letterpaper]{article}

\usepackage[pagenumbers]{iccv} 

\usepackage{multirow} 
\usepackage[table,xcdraw]{xcolor}
\definecolor{mygray}{gray}{.875}

\def\purple{\textcolor[RGB]{220, 0, 220}}
\def\blue{\textcolor[RGB]{80,120,200}}
\definecolor{myorange}{HTML}{FFEEB4}
\usepackage{pifont}
\usepackage{fontawesome}

\definecolor{iccvblue}{rgb}{0.21,0.49,0.74}
\usepackage[pagebackref,breaklinks,colorlinks,allcolors=iccvblue]{hyperref}

\def\paperID{1062} 
\def\confName{ICCV}
\def\confYear{2025}

\title{Consistent-Point: Consistent Pseudo-Points for Semi-Supervised \\ Crowd Counting and Localization}

\author{Yuda Zou, Zelong Liu, Yuliang Gu, ~Bo Du, ~Yongchao Xu\textsuperscript{\faEnvelope}\\
School of Computer Science, Wuhan University, Wuhan, China\\
{\tt\small zouyuda@whu.edu.cn yongchao.xu@whu.edu.cn}
}

\begin{document}
\maketitle
\begin{abstract}
Crowd counting and localization are important in applications such as public security and traffic management. Existing methods have achieved impressive results thanks to extensive laborious annotations. This paper propose a novel point-localization-based semi-supervised crowd counting and localization method termed \textit{Consistent-Point}.
We identify and address two inconsistencies of pseudo-points, which have not been adequately explored. 
To enhance their position consistency, we aggregate the positions of neighboring auxiliary proposal-points.
Additionally, an instance-wise uncertainty calibration is proposed to improve the class consistency of pseudo-points. 
By generating more consistent pseudo-points, Consistent-Point provides more stable supervision to the training process, yielding improved results.
Extensive experiments across five widely used datasets and three different labeled ratio settings demonstrate that our method achieves state-of-the-art performance in crowd localization while also attaining impressive crowd counting results. 
The code will be available.
\end{abstract}    
\section{Introduction}
\label{sec:introduction}

\begin{figure}[ht]
  \centering
\includegraphics[width=.95\linewidth]{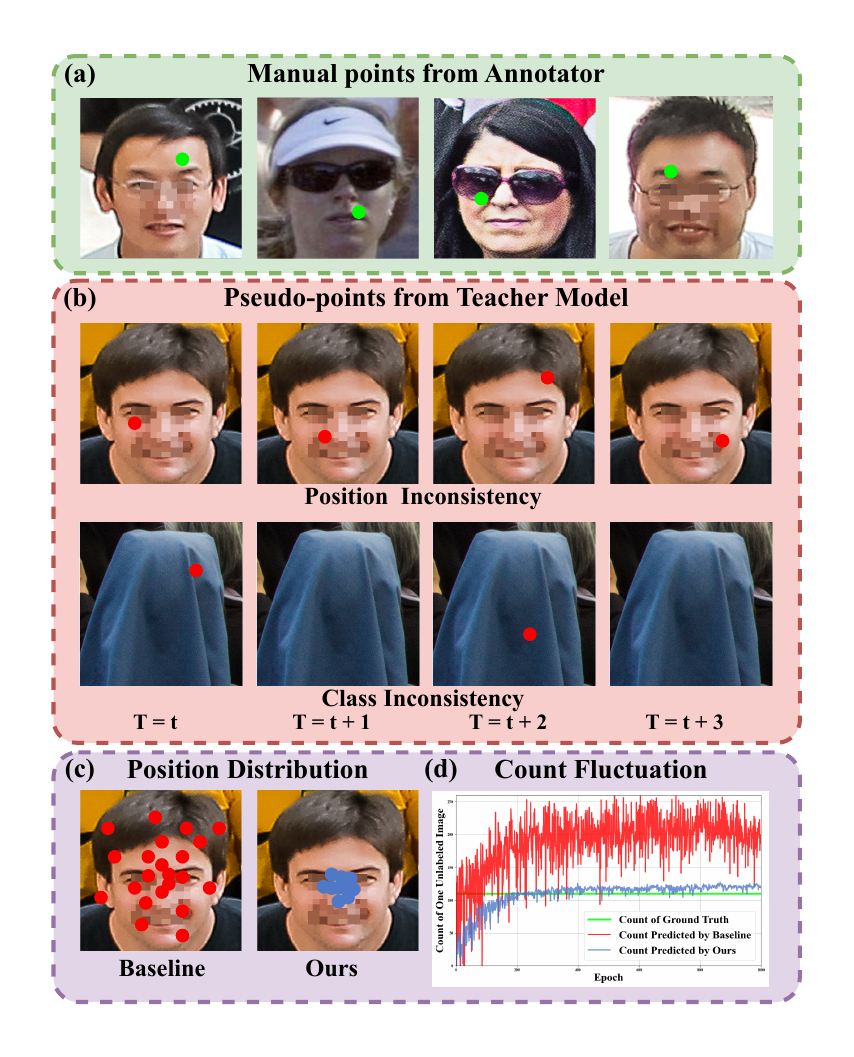}
  \caption{Annotator subjectivity introduces position inconsistency for manual annotation points. The position and classification inconsistency of pseudo-points arise from the regression branch and classification branch of the point-localization teacher model with changing parameters in the training process, respectively. 
  }
  \label{fig: motivation}
\end{figure}

Crowd counting, which aims to count people in an image, has growing practical value in areas such as public security and traffic management~\cite{liang2022transcrowd, iterative_prototype_adaptation_2023_ICCV, interactive_class_agnostic_2023_ICCV, han2022dr, hui2024class, zhu2024zero}. 
Fully-supervised methods~\cite{ma2021learning, abousamra2021localization_chenchao, liu2019context_can, han2023steerer_STEERER, huang2023bad_weather, Pyramid_p2p} have achieved promising results. 
These methods can generally be divided into two categories: density-based ones~\cite{wan2020modeling_NoiseCC_nips, wan2021fine, sun2023indiscernible_underwater, zhangqi2019wide, zhangqi2021cross, ranasinghe2024crowddiff, peng2024single, guo2024regressor} and localization-based ones. Density-based methods leverage regression techniques to estimate the density distribution of people within the image. In contrast, localization-based methods~\cite{sam2020locate_box, song2021rethinking_P2P, liu2023point_PET, liang2022CLTR, Pyramid_p2p}, focus on achieving counting by locating each individual through bounding boxes~\cite{sam2020locate_box} or 2D points~\cite{song2021rethinking_P2P, liu2023point_PET, liang2022CLTR, Pyramid_p2p}, which directly provide exact localization information and offer greater practical potential.

These fully-supervised methods typically require large amounts of annotated data to achieve good performance. However, annotating a single crowd image often involves marking hundreds or even thousands of person heads with 2D points, which is time-consuming and labor-intensive~\cite{liang2023crowdclip, knobel2024learning_without_annotations, pelhan2024dave, d2024afreeca}. 
Compared with dense annotations, unlabeled crowd images can be obtained cheaply and conveniently~\cite{wan2024robust}.
To alleviate the annotation burden, some semi-supervised crowd counting methods~\cite{liu2019exploiting_rank_TPAMI, sindagi2020learning_GP} have been proposed.  
These methods leverage unlabeled images to extract valuable information, thereby improving the final accuracy.

Similar to fully-supervised methods, semi-supervised crowd counting methods can also be divided into density-based ones and localization-based ones. 
Some existing density-based semi-supervised methods capture counting information from unlabeled crowd images by designing auxiliary tasks. 
For example, IRAST~\cite{liu2020semi_IRAST}, DACount~\cite{lin2022semi_DACount}, and P$^3$Net~\cite{HuiLin_pami_P3} introduce pixel-wise or patch-wise density classification tasks to enhance the density perception of the network.
Recently, mean-teacher~\cite{tarvainen2017mean_teacher} emerges as the mainstream for semi-supervised crowd counting~\cite{meng2021spatial_SUA, wang2023semi_STC-Crowd, li2023calibrating_CU, lin2023optimal_OT-M, scale-based_active_learning}. 
Pseudo-labels predicted by the teacher model (pseudo-density-map for density-based methods and pseudo-point for point-localization-based method) on unlabeled images serve as supervision for the student model training. The intrinsic noise in the pseudo labels degrades the training of the student model. To mitigate this, some density-based methods~\cite{meng2021spatial_SUA, wang2023semi_STC-Crowd} incorporate pixel-wise uncertainty into the predicted pseudo-density-map. 
Departing from the density-based paradigm, some mean-teacher methods leverage a typical point-localization-based crowd counting method~\cite{song2021rethinking_P2P} as the counting baseline~\cite{li2023calibrating_CU, scale-based_active_learning}.
CU~\cite{li2023calibrating_CU} estimates patch-wise uncertainty to select reliable unlabeled patches for training the student model, and SAL~\cite{scale-based_active_learning} utilizes active learning to select informative images as labeled split.  
Although existing localization-based mean-teacher methods have achieved impressive results, they may overlook the inconsistency of pseudo-points across different training times. 
Such inconsistencies introduce unstable supervision signals, which confuse the model training and ultimately degrade its performance.

In fact, due to the subjectivity of annotators, position inconsistency also exists in manually annotated points (Fig.~\ref{fig: motivation}a) between different heads, resulting in confusing the training and degraded counting performance~\cite{zou2024_SAE}. 
In fully-supervised crowd counting, BL~\cite{ma2019bayesian_BL} and NoiseCC~\cite{wan2023modeling_NoiseCC_Tpami} address this by designing a Gaussian-based loss for density-based counting methods. MAN~\cite{lin2022boosting_MAN} mitigates the impact of annotation errors by ignoring the top 10\% of their losses.
Different from manually annotated points, the pseudo-points, which are predicted by the teacher model with continuously changing parameters during the training process, may suffer from more severe inconsistency issues from two aspects. 
Generally, the point-localization-based method consists of one regression branch for localization and one classification branch to classify.
On one hand, parameter changes in the regression branch can lead to position inconsistency to pseudo-points (first row of Fig.~\ref{fig: motivation}b) for the same head. 
On the other hand, the classification branch with varying parameters leads to ambiguous instances being inconsistently predicted as either a person at some times or not a person at other times (second row of Fig.~\ref{fig: motivation}b).
Consequently, these inconsistency issues may be more critical but have not been well explored in localization-based semi-supervised crowd counting.

Building upon the mean teacher paradigm, we introduce a novel semi-supervised crowd counting approach termed \textit{Consistent-Point}, utilizing the point-localization-based P2PNet~\cite{song2021rethinking_P2P} as both the teacher and student models.
To address the position inconsistency from the regression branch and the class inconsistency from the classification branch in the teacher P2PNet, we design a Position Aggregation (PA) module and an Instance-wise Uncertainty Calibration (IUC) module, respectively.
For each pseudo-point, the PA takes the neighboring four auxiliary proposal-points and aggregate their positions, resulting in the position-consistent pseudo-point. 
Smoothed by the auxiliary points, this position-consistent point presents more position consistency (Fig.~\ref{fig: motivation}c).
As for the class inconsistency, the IUC module weights pseudo-points with their inherent classification scores predicted from the teacher classification branch. 
This helps to mitigate class inconsistency by reducing the weight of pseudo-points of those ambiguous instances, resulting in more stable and accurate count results of unlabeled images (Fig.~\ref{fig: motivation}d).
The main contributions of this work are threefold: 
\begin{itemize}
\item We analyze the inconsistency of pseudo-points for unlabeled images and propose a novel localization-based semi-supervised crowd counting method to improve their consistency effectively. 

\item We propose two simple yet effective modules, PA and IUC, to improve pseudo-point consistency in terms of position and class, respectively.

\item The proposed method achieves state-of-the-art crowd localization across 15 different settings, covering five common datasets with three annotation ratios. For crowd counting, our method also achieves impressive results.

\end{itemize}

\section{Related work}
\label{sec: related work}

\subsection{Fully-Supervised Crowd Counting}
\label{subsec: related work fully-supervised crowd counting}
There are mainly two types of crowd counting methods: density-based methods
and localization-based methods.
Density-based ones~\cite{huang2023counting_AWCC, rsi_cheng2022rethinking, wan2019adaptive_density, wan2020kernel_based, li2018csrnet, wang2020distribution, ma2020learning_DPN-IPSM, mo2024countformer} regard the counting task as a density map prediction task.
To tackle the scale variation problem arising from perspective~\cite{wu2019one_perspective}, some new counting network architectures~\cite{xu2019learn_autoscale_iccv, lin2022boosting_MAN} are proposed.
For instance, MAN~\cite{lin2022boosting_MAN} proposed a transformer-based learnable local region attention to address the problem of large-scale variation in crowd images. 
In addition to the scale variation problem, some works focus on the location inconsistency issue in manual annotation points. 
Instead of original hard pixel-wise MSE Loss, BL~\cite{ma2019bayesian_BL} and NoiseCC~\cite{wan2020modeling_NoiseCC_nips} alleviate the location inconsistency by designing Gaussian distance-based expectation losses.

Although density-based methods show promising results in crowd counting, they often struggle to accurately localize individuals.
In contrast, localization-based counting methods\cite{sam2020locate_box, liu2023point_PET, song2021rethinking_P2P} achieve counting by locating each individual separately, offering potential for broader applications. Sam \textit{et al.}~\cite{sam2020locate_box} regard the counting task as object detection using pseudo bounding boxes, but encounters challenges in dense areas. To tackle this, some works~\cite{song2021rethinking_P2P, liu2023point_PET, APGCC} discard the bounding box and instead directly utilize annotation points for individual localization.

\begin{figure*}[t]
  \centering
  \includegraphics[width=.92\linewidth]{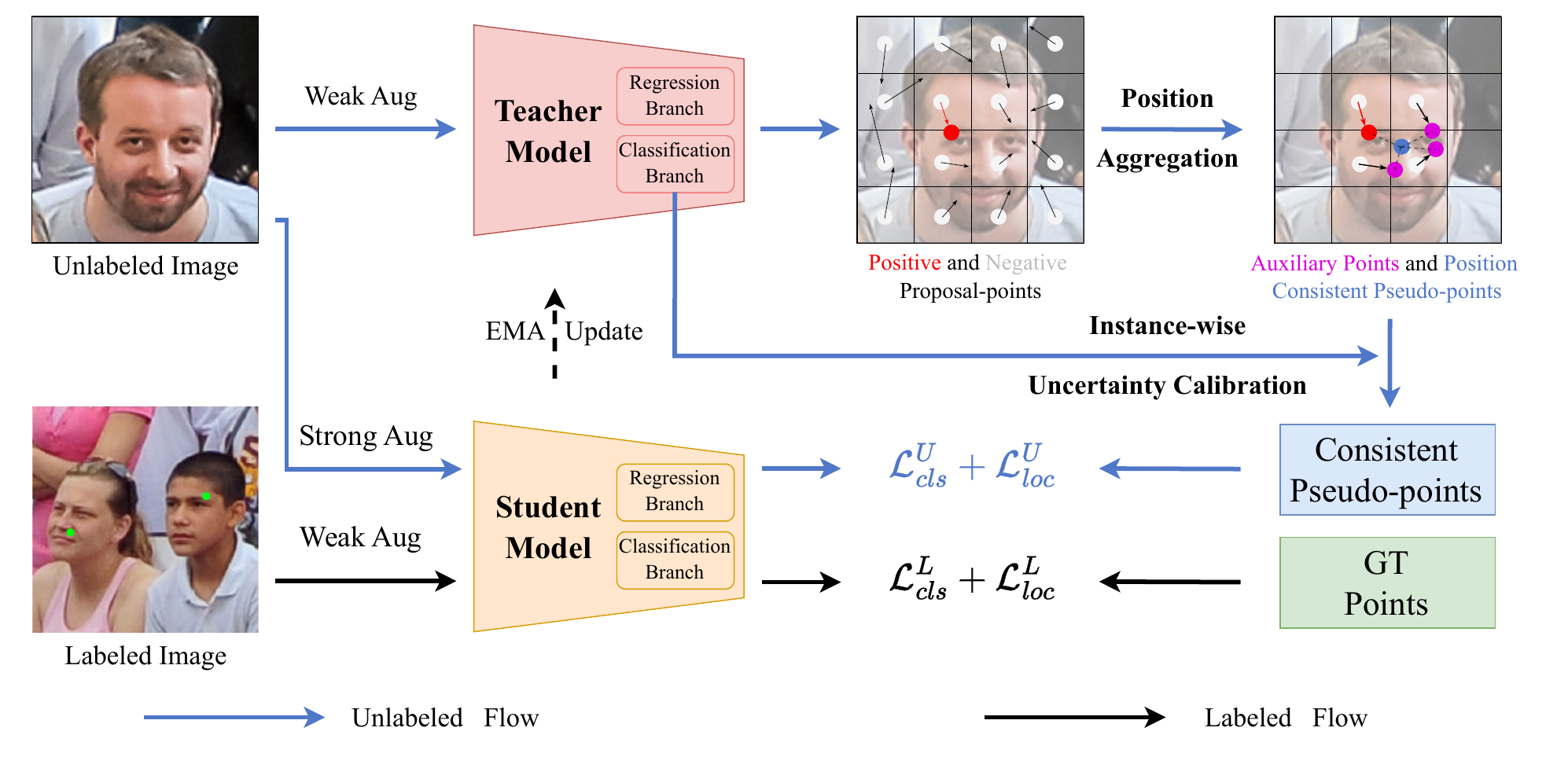}
  \caption{The pipeline of \textit{Consistent-Point}. We design two modules for point-localization-based crowd counting under the mean-teacher paradigm. Position Aggregation (PA) alleviates the position inconsistency from the regression branch; Instance-wise Uncertainty Calibration (IUC) alleviates the class inconsistency arising from the classification branch.  
  }
  \label{fig: pipeline}
    
\end{figure*}

\subsection{Semi-Supervised Density-Based Counting}
\label{subsec: related_work_limited_annotations}
Annotation of images for crowd counting tasks is time-consuming and labor-intensive. To alleviate annotation costs, an increasing number of works~\cite{sindagi2020learning_GP, liu2020semi_IRAST, chen2023multi} focus on semi-supervised crowd counting, which leverages unlabeled images to enhance the counting accuracy of models. 
Some methods extract information from unlabeled data by designing auxiliary tasks. 
For example, Liu et al.~\cite{liu2018leveraging_rank_cvpr, liu2019exploiting_rank_TPAMI} propose a learning-to-rank (L2R) module to consider the inherent relationship between people numbers within an unlabeled image patch and its sub-patch.
DACount~\cite{lin2022semi_DACount} and P$^3$Net~\cite{HuiLin_pami_P3} enhance the model's ability to recognize  density levels by assigning distinct different labels to image patches.
Recently, several methods adopt the mean-teacher paradigm, which utilizes pseudo labels predicted by the teacher model to supervise the training of the student model, and achieve impressive performance. 
The key is how to obtain high-quality pseudo labels.
Some density-based methods~\cite{meng2021spatial_SUA, wang2023semi_STC-Crowd} reduce noise in pseudo-density-maps by introducing pixel-wise uncertainty.

\subsection{Semi-Supervised Localization-Based Counting}
\label{subsec: related_work_semi-supervised localization-based Counting}
Existing semi-supervised localization-based methods~\cite{lin2023optimal_OT-M, li2023calibrating_CU, scale-based_active_learning} adopt the mean-teacher paradigm, with pseudo-labels represented as 2D points. 
OT-M~\cite{lin2023optimal_OT-M} employs the optimal transport algorithm to obtain pseudo-points from the predicted map.
Building upon the point-localization-based model P2PNet~\cite{song2021rethinking_P2P}, CU~\cite{li2023calibrating_CU} employs patch-wise uncertainty to select patches containing high-quality pseudo-points for student model training.
Unlike traditional semi-supervised methods, which use the same partitioning for labeled and unlabeled data, SAL~\cite{scale-based_active_learning} adopts active learning to select challenging images as labeled data and easy ones as unlabeled data.
While pseudo-point-based semi-supervised crowd counting methods have shown promising results, the inconsistency of pseudo-points is often ignored. 
This issue confuses the student training, thereby reducing counting accuracy.
To address this, we propose a point-localization-based semi-supervised crowd counting method, \textit{Consistent-Point}, which mitigates both position and class inconsistencies in pseudo-points.
Consistent pseudo-points can provide more stable supervision for training, resulting in improved crowd counting and localization accuracies.

\section{Preliminary on Point-Localization-Based Crowd Counting}
\label{sec:preliminary}

Following CU~\cite{li2023calibrating_CU} and SAL~\cite{scale-based_active_learning}, we adopt P2PNet~\cite{song2021rethinking_P2P}, the typical point-localization crowd counting method, as our teacher model and student model. It consists of three main components: point-proposal prediction, proposal-target point matching, and loss function.

\noindent\textbf{Point Proposal Prediction} generates point proposals from the feature map $\mathcal{F}$ output by the backbone network. Following $\mathcal{F}$, P2PNet adopts two parallel branches: a regression branch for predicting position offsets for each proposal-point, $\hat{p}_j = (\hat{x}_j, \hat{y}_j)$, and a classification branch for determining their classes by predicting classification scores, $\hat{c}_j \in [0, 1]$.

\noindent\textbf{Proposal-Target Point Matching} adopts the one-to-one Hungarian matching algorithm~\cite{kuhn1955hungarian} to conduct the matching operation $\Omega(\mathcal{P}, \hat{\mathcal{P}}, \mathcal{D})$, where each ground truth target point from $\mathcal{P}$ is matched to a proposal-point from $\hat{\mathcal{P}}$.
The matching process is guided by the pair-wise cost matrix $\mathcal{D}$, with dimensions $N \times M$, where $N$ and $M$ represent the numbers of ground truth points and predicted points, respectively.
After the matching process $\xi$, each ground truth point $p_i$ is optimally paired with a proposal-point $\hat{p}_{\xi(i)}$.
The set of matched proposal-points, $\hat{\mathcal{P}}_{pos} = \{\hat{p}_{\xi(i)} | i \in \{1, \ldots, N\}\}$, are considered positive, while the remaining unmatched ones, $\hat{\mathcal{P}}_{neg} = \{\hat{p}_{\xi(i)} | i \in \{N + 1, \ldots, M\}\}$, are negative.

\noindent\textbf{Loss Function} consists of Euclidean loss $\mathcal{L}^L_{loc}$ for localization of positive proposal-points and Cross Entropy loss $\mathcal{L}^L_{cls}$ for classification of all the proposal-points. The combined loss $\mathcal{L}^L$ is defined as:
\begin{equation}
\mathcal{L}^L_{loc} = \frac{1}{N} \sum_{i=1}^{N} \|\mathbf{p}_i - \hat{\mathbf{p}}_{\xi(i)}\|_2^2,
\label{eq: labeled localization loss}
\end{equation}
\begin{equation}
\mathcal{L}^{L}_{cls} = -\frac{1}{M}\left(\sum_{i=1}^{N}\log\hat{c}_{\xi(i)} + \lambda_1 \sum_{i=N+1}^{M}\log(1 - \hat{c}_{\xi(i)})\right),
\label{eq: labeled classification loss}
\end{equation}
\begin{equation}
\mathcal{L}^L = \mathcal{L}^{L}_{cls} + \lambda_2\mathcal{L}^{L}_{loc},
\label{eq: labeled loss}
\end{equation}
where $\hat{c}_{\xi(i)}$ is the classification score of the matched proposal-point $\hat{p}_{\xi(i)}$.  
Both $\lambda_1$ and $\lambda_2$ are hyperparameters introduced in the P2PNet~\cite{song2021rethinking_P2P}, and we retain their values ($\lambda_1=0.5, \lambda_2=2e^{-4}$) as proposed throughout this work.

\section{Methodology}
\label{sec:methodology}

\subsection{Overview}
\label{subsec:methodology_overview}

In the field of semi-supervised crowd counting and localization, the mean-teacher paradigm~\cite{tarvainen2017mean_teacher} has become the dominant approach~\cite{li2023calibrating_CU, wang2023semi_STC-Crowd, scale-based_active_learning}. This paradigm typically involves a teacher model and a student model with identical architecture. The student model is trained not only on labeled images with annotated labels but also on unlabeled images, supervised by pseudo-labels predicted by the teacher model. The teacher model's parameters are then updated as an exponential moving average (EMA) of the student model's parameters.
However, due to the constant updates in the model parameters, the pseudo-labels generated by the teacher model become unstable during training. For density-based models, STC-Crowd~\cite{wang2023semi_STC-Crowd} mitigates this instability in pseudo-density-maps by incorporating temporal ensembling~\cite{feng2021temporal_consistency} and pixel-wise uncertainty. In contrast, the issue has not been explored in point-localization-based models, where pseudo-labels are represented as 2D points. As these pseudo-points are predicted from both the regression and classification branches of the teacher model, their inconsistency during training primarily arises from the inconsistent predictions of these two branches at different training times.

To alleviate these, we propose a novel point-localization-based semi-supervised crowd counting method, termed \textit{Consistent-Point}. Fig.~\ref{fig: pipeline} illustrates an overview of our framework. 
Following CU~\cite{li2023calibrating_CU} and SAL~\cite{scale-based_active_learning}, both the teacher and student models employ P2PNet~\cite{song2021rethinking_P2P}, a typical point-localization-based network with a regression branch and a classification branch for point localization and classification, respectively. We introduce two modules: Position Aggregation (PA) and Instance-wise Uncertainty Calibration (IUC), designed to address the position inconsistency from the regression branch and the classification inconsistency from the classification branch, respectively.

\subsection{Position Aggregation}
\label{subsec: position aggregation}

The quality of annotations significantly impacts model training and is critical to the final performance. In crowd counting and localization, annotations are typically provided as 2D points marking the location of each individual's head. However, the subjective nature of annotators' selections often results in inconsistencies in the positioning of these manual annotation points across different heads (Fig.~\ref{fig: motivation}a). This inconsistency confuses the training of fully-supervised models, thereby degrading counting and localization accuracy~\cite{zou2024_SAE, wan2023modeling_NoiseCC_Tpami, ma2019bayesian_BL}.
In contrast, in point-localization-based semi-supervised crowd counting, the positions of pseudo-points, $\hat{\mathcal{P}}^t = \{\hat{p}^t_i | i \in \{1, \ldots, N^U\}\}$, of one unlabeled image are predicted by the regression branch of the teacher P2PNet, which has continuously changing parameters during the training process. 
Consequently, pseudo-points suffer from position inconsistency issue even within the same heads during the training (top row of ($b$) part in Fig.~\ref{fig: motivation}).

In fact, due to the one-to-one matching strategy of P2PNet, a pseudo-point (\ie, a positive proposal-point) suppresses its neighboring proposal-points and they will be classified as negative in the classification branch. 
However, owing to the high similarity in features, the regression branch can provide similar localization results for the psuedo-point and its neighboring proposal-points.
Without loss of generality, we assume that the localization of a neighecooring point follows a Gaussian distribution. Consequently, the similar localization of $K$ points, when averaged, will also follow a Gaussian distribution, but with a reduced variance:
\begin{equation}
p^{aux} \in N(0, \sigma^2)~,
\end{equation}
\begin{equation}
\hat{cp}^t = \frac{1}{K}\sum\limits_{}^{K} p^{aux} \in N(0, \frac{1}{K}\sigma^2)~.
\label{eq: gaussian distribution}
\end{equation}
Therefore, we propose Position Aggregation (PA) module to improve the position consistency of pesudo-points with the help of the neighboring proposal-points. Specifically, for each pseudo-point (\eg, \textcolor{red}{red point} in Fig.~\ref{fig: pipeline}), PA first refers to its neighboring four proposal-points as the auxiliary points (\eg, \purple{purple point} in Fig.~\ref{fig: pipeline}) and aggregates their position by simply averaging, resulting in the position-consistent pseudo-point.
Smoothed by the auxiliary points, the position-consistent pseudo-points (\eg, \blue{blue points} in Fig.~\ref{fig: pipeline}), $\hat{\mathcal{CP}}^t = \{\hat{cp}^t_i | i \in \{1, \ldots, N^U\}\}$, are stable with more position consistency (Fig.~\ref{fig: motivation}c) compared with the initial pseudo-points.
The notation $^t$ indicates it is predicted from the teacher model.
utilizeing these position-consistent pseudo-points to supervise the training results in better crowd counting and localization performance.

\subsection{Instance-wise Uncertainty Calibration}
\label{subsec: instance-wise uncertainty calibration}

Besides the position inconsistency introduced by the regression branch, the classification branch bring about class inconsistency. As illustrated in the second row of Fig.~\ref{fig: motivation}b, the teacher P2PNet may oscillate between classifying an ambiguous instance as either a person or background. This inconsistency occurs due to the continuously changing parameters of the classification branch during training. As a result, the instance is sometimes predicted as a person and assigned a pseudo-point (a positive proposal-point), while at other times it is assigned as background with no pseudo-point. Directly using such inconsistent pseudo-points as supervision signals can confuse the student P2PNet training provess, ultimately degrading its crowd counting and localization performance.

To address this, we propose a simple instance-wise uncertainty calibration (IUC) to mitigate the impact of such frequent class changes. Specifically, as each pseudo-point is a proposal-point classified as positive by the classification branch of the teacher model, we assign weights to these position-consistent pseudo-points, $\hat{\mathcal{CP}}^t = \{\hat{cp}^t_i | i \in \{1, \ldots, N^U\}\}$, based on their own inherent classification scores, $\hat{\mathcal{C}}^t = \{\hat{c}^t_i | i \in \{1, \ldots, N^U\}\}$, where $\hat{c}^t_i \in [0.5, 1]$.
With the set of matched proposal-points, $\hat{\mathcal{P}}_{pos}^s = \{\hat{p}_{\xi^t(i)}^s | i \in \{1, \ldots, N^U\}\}$, and the unmatched ones, $\hat{\mathcal{P}}_{neg}^s = \{\hat{p}_{\xi^t(i)}^s | i \in \{N^U + 1, \ldots, M^U\}\}$, of student model, 
the classification loss and the localization loss for the unlabeled image are defined as:
\begin{equation}
\mathcal{L}^U_{loc} = \frac{1}{N^U} \sum_{i=1}^{N^U} \|\hat{\mathbf{cp}}^t_i - \hat{\mathbf{p}}^s_{\xi^t(i)}\|_2^2,
\label{eq: unlabeled localization loss}
\end{equation}
\begin{equation}
\mathcal{L}^U_{cls} = \frac{-1}{M^U}\left(\sum_{i=1}^{N^U}w_i\log\hat{c}^s_{\xi^t(i)} + \lambda_1 \sum_{i=N^U+1}^{M^U}\log(1 - \hat{c}^s_{\xi^t(i)})\right)
\label{eq: unlabeled classification loss}
\end{equation}
\begin{equation}
w_i = \frac{\hat{c}^t_i-0.5}{0.5},
\label{eq: weight}
\end{equation}
where $\hat{\mathbf{p}}_{\xi^t(i)}$ is the student-model-predicted proposal-point matched to the teacher-model-predicted position-consistent pseudo-point $\hat{cp}^t_i$. $\hat{c}^s_{\xi^t(i)}$ is the classification score after softmax operation of the matched proposal-point $\hat{p}^s_{\xi^t(i)}$.
The instance-wise weight $w_i$ strengthens pseudo-points with low class inconsistency while weakening those associated with ambiguous instances that exhibit high class inconsistency, resulting in the final consistent pseudo-points. 
These consistent pseudo-points can provide the student P2PNet with more stable and higher-quality training guides, thereby boosting the final counting and localization performance.

\subsection{Semi-Supervised Crowd Localization}
\label{subsec:methodologt_semi-supervised_crowd_localization}
Throughout the training process, the student P2PNet is trained using both labeled data $D^L$ with corresponding manual annotation points $\mathcal{P}$ and unlabeled data $D^U$ using the consistent pseudo-points $\mathcal{P}^c$ from the teacher P2PNet.
For the labeled data $D^L$, the loss function is defined in Eq.~\eqref{eq: labeled loss}.
Based on Eq.~\eqref{eq: unlabeled localization loss} and Eq.~\eqref{eq: unlabeled classification loss}, 
the combined loss function for the unlabeled image is given by: 
\begin{equation}
\mathcal{L}^U = \mathcal{L}^{U}_{cls} + \lambda_2\mathcal{L}^{U}_{loc}~~.
\label{eq: unlabeled loss}
\end{equation}
The overall loss function for the student P2PNet is given by
\begin{equation}
\mathcal{L} = \mathcal{L}^L + \lambda \mathcal{L}^U,
\label{eq:loss}
\end{equation}
where $\lambda$ is used to balance the losses for labeled and unlabeled data, which is the only tuned hyperparameter and set to $0.1$ empirically. 
The teacher P2PNet is updated with the Exponential Moving Average of the student P2PNet.

\section{Experiments}
\label{sec: experimentation}

\subsection{Experimental Setting}
\label{subsec: experimental_setting}

\noindent
\textbf{Datasets.} we validate the effectiveness of our method on five widely used crowd counting and localization datasets: ShanghaiTech Part A~\cite{zhang2016single_SH_MCNN}, ShanghaiTech Part B~\cite{zhang2016single_SH_MCNN}, UCF-QNRF~\cite{idrees2018composition_UCF-QNRF}, JHU-Crowd++~\cite{sindagi2020jhu} and NWPU-Crowd~\cite{wang2020nwpu}.

\begin{table*}[t]
    \centering

        \resizebox{1.\linewidth}{!}{
		\begin{tabular}{l | c| |c c c| c c c|c c c|c c c | c c c}
			\specialrule{0.2em}{0pt}{0pt}
   
\rowcolor{mygray}			  & Labeled & \multicolumn{3}{c|}{SHA}  & \multicolumn{3}{c|}{SHB} & \multicolumn{3}{c|}{UCF-QNRF} & \multicolumn{3}{c|}{JHU-Crowd++} & \multicolumn{3}{c}{NWPU-Crowd}  \\ 
            \cline{3-17}
\rowcolor{mygray}	\multirow{-2}{*}{Method}		& Ratio &  F(\%)$\uparrow$ & P(\%)$\uparrow$ & R(\%)$\uparrow$&  F(\%)$\uparrow$ & P(\%)$\uparrow$ & R(\%)$\uparrow$&  F(\%)$\uparrow$ & P(\%)$\uparrow$ & R(\%)$\uparrow$&  F(\%)$\uparrow$ & P(\%)$\uparrow$ & R(\%)$\uparrow$ & F(\%)$\uparrow$ & P(\%)$\uparrow$ & R(\%)$\uparrow$\\
			 \specialrule{0.2em}{0pt}{0pt}

\multicolumn{17}{c}{\textbf{(a) large threshold $\boldsymbol{\sigma_l = 8}$ }} \\

 \specialrule{0.2em}{0pt}{0pt}
        OT-M~\cite{lin2023optimal_OT-M}   & 5\%            &65.8 &67.2 & 64.6  &74.3 & 74.8 &73.8 &68.8 &69.4 & 68.2 &66.6 &66.5 & 66.7 &59.8 &64.9 & 55.5\\
        CU~\cite{li2023calibrating_CU}         & 5\%     &67.8 &68.9 & 66.8  &75.0 &75.2 & 74.8 &70.5 &70.9 & 70.1 &66.2 &\textbf{70.1} & 62.7 &61.0 &62.0 & 60.1 \\

\rowcolor{myorange} \textbf{Ours}    & 5\%        &\textbf{71.5} &\textbf{70.7} & \textbf{72.3}  &\textbf{77.2} &\textbf{77.4} & \textbf{76.9} &\textbf{72.2} &\textbf{73.5} & \textbf{71.0} &\textbf{68.1} & 68.7 & \textbf{67.5} &\textbf{70.2} &\textbf{69.0} & \textbf{71.4}    \\
			 \specialrule{0.1em}{0pt}{0pt}
        OT-M~\cite{lin2023optimal_OT-M}   & 10\%      &69.7 &70.9 & 68.5  &76.2 &76.6 & 75.7 &70.1 &70.4 & 69.9 &67.8 &70.2 & 65.6 &62.2 &65.7 & 59.1 \\

        CU~\cite{li2023calibrating_CU}       & 10\%  & 68.5 & 70.7 &  66.5    &76.3 &76.6 & 76.0 &72.8 &73.2 & 72.5  &67.8 &70.9 & 65.0 &66.2 &66.3 & 66.1    \\
        SAL~\cite{scale-based_active_learning}         & 10\%     &71.2 &73.8 & 68.7  &76.9 &77.3 & 76.5 &73.3 &72.9 & 73.8 &68.6 &69.2 & 68.1 &66.6 &70.1 & 63.5\\

\rowcolor{myorange}               \textbf{Ours}   & 10\% &\textbf{73.8} &\textbf{75.3} & \textbf{72.3}  &\textbf{78.6} &\textbf{80.2} & \textbf{77.2} & \textbf{75.5} &\textbf{76.4} &\textbf{74.5} &\textbf{69.7} &\textbf{71.2} & \textbf{68.2}  &\textbf{73.8} &\textbf{73.3} & \textbf{74.3} \\

			 \specialrule{0.1em}{0pt}{0pt}
        OT-M~\cite{lin2023optimal_OT-M}   & 40\%     &71.4 &73.1 & 69.8  &78.4 &79.1 & 77.7 &73.9 &73.6 & 74.2 &68.6 &68.9 & 68.3 &68.3 &68.1 & 68.5\\

        CU~\cite{li2023calibrating_CU}           & 40\%  &74.2 &74.2 & 74.2  &79.3 &79.9 & 78.8 &75.5 &76.4 & 74.6 &72.5 &74.6 & 70.5 &68.6 &70.5 & 66.8\\
        SAL~\cite{scale-based_active_learning}         & 40\%   &75.8 &77.1 & 74.6   &80.0 &80.6 & 79.5 &74.7 &75.2 & 74.3 &73.0 &73.6 & 72.4 &71.0 &71.2 & 70.8\\

\rowcolor{myorange}               \textbf{Ours}   & 40\%      &\textbf{78.1} &\textbf{78.5} & \textbf{77.6}  &\textbf{82.9} &\textbf{82.8} & \textbf{83.0} &\textbf{78.1} &\textbf{78.8} & \textbf{77.4} &\textbf{74.1} &\textbf{74.8} & \textbf{73.4} &\textbf{78.2} &\textbf{77.9} & \textbf{78.5}   \\

			 \specialrule{0.2em}{0pt}{0pt}

            \multicolumn{17}{c}{\textbf{(b) small threshold $\boldsymbol{\sigma_s = 4}$ }} \\
            
\specialrule{0.2em}{0pt}{0pt}

OT-M~\cite{lin2023optimal_OT-M}   & 5\%           &38.0 &38.7 & 37.2  &49.2 &48.8 & 49.6 &38.1 &38.5 & 37.8 &41.5 &41.4 & 41.5  &34.8 &37.8 & 32.3 \\
CU~\cite{li2023calibrating_CU}         & 5\%     &38.3 &38.9 & 37.7  &50.0 &49.2 & 50.8 &40.9 &41.2 & 40.7 &40.5 &41.6 & 39.4  &36.5 &37.0 & 35.9\\

\rowcolor{myorange}    \textbf{Ours}    & 5\%            &\textbf{42.0} &\textbf{41.5} &\textbf{42.4}  &\textbf{51.1} &\textbf{51.2} & \textbf{50.9} &\textbf{44.4} &\textbf{45.2} & \textbf{43.6} &\textbf{42.2} &\textbf{42.6} & \textbf{41.8} & \textbf{48.0} & \textbf{47.2} & \textbf{48.9} \\

             \specialrule{0.1em}{0pt}{0pt}
        OT-M~\cite{lin2023optimal_OT-M}   & 10\%         &40.5 &41.3 &39.8  &49.6 &49.8 & 49.3 &42.5 &41.9 & 43.2 &42.7 &44.2 & 41.3 &36.4 &38.5 & 34.6\\

        CU~\cite{li2023calibrating_CU}       & 10\%  & 39.3 & 40.6 &  38.2    &49.9 &50.1 & 49.8 &46.6 &46.9 & 46.4 &42.7 &44.6 & 40.9&41.4 &41.4 & 41.3  \\
        SAL~\cite{scale-based_active_learning}         & 10\%    &41.3 & 42.8 & 39.9  &50.2 &49.2 & 51.3 &47.2 &47.6 & 46.9 &43.6 &44.9 & 42.3 &41.5 &43.7 & 39.6 \\

\rowcolor{myorange}        \textbf{Ours}   & 10\%    &\textbf{45.8} &\textbf{46.8} & \textbf{44.9}  &\textbf{52.6} &\textbf{53.6} & \textbf{51.6} &\textbf{49.9} &\textbf{50.5} & \textbf{49.3} &\textbf{45.7} &\textbf{46.7} & \textbf{44.7} &\textbf{52.9} &\textbf{52.6} & \textbf{53.3} \\

\specialrule{0.1em}{0pt}{0pt}
OT-M~\cite{lin2023optimal_OT-M}   & 40\%        &43.7 &44.8 & 42.7  &52.9 &53.3 &52.4 &48.0 &47.9 & 48.2 &44.8 &44.3 & 45.3 &43.6 &43.5 & 43.7 \\

CU~\cite{li2023calibrating_CU}           & 40\%    &46.3 &46.3 & 46.3  &55.3 &55.7 & 54.9 &49.9 &50.5 & 49.3 &50.9 &52.4 & 49.5 &43.2 &44.3 & 42.0 \\
SAL~\cite{scale-based_active_learning}         & 40\%    &47.7 &48.5 & 46.9  &56.8 &57.3 & 56.3 &50.7 &51.2 & 50.3  &51.3 & 51.8 &50.8 &46.9 &46.7 & 47.2 \\

\rowcolor{myorange}    \textbf{Ours}   & 40\%       &\textbf{52.9} &\textbf{53.3} & \textbf{52.6}  &\textbf{60.0} &\textbf{59.9} & \textbf{60.0}   &\textbf{54.7}  &\textbf{55.2}  & \textbf{54.2} &\textbf{53.0} &\textbf{53.5} & \textbf{52.5}  &\textbf{57.3} &\textbf{57.0} & \textbf{57.5} \\

    	\specialrule{0.2em}{0pt}{0pt}
	    \end{tabular}
        }
	\caption{Quantitative crowd localization comparison with state-of-the-art semi-supervised crowd localization methods on ShanghaiTech Part A dataset, Part B dataset~\cite{zhang2016single_SH_MCNN}, UCF-QNRF dataset~\cite{idrees2018composition_UCF-QNRF}, JHU-Crowd++ dataset~\cite{sindagi2020jhu}, and NWPU-Crowd dataset~\cite{wang2020nwpu} under large threshold $\sigma_l = 8$ and small threshold $\sigma_s = 4$, respectively. 
 The best performance is in \textbf{bold}.
 }
\label{table: localization results under large threshold}

\end{table*}

\begin{figure*}[ht]
  \centering
  \includegraphics[width=1.\linewidth]{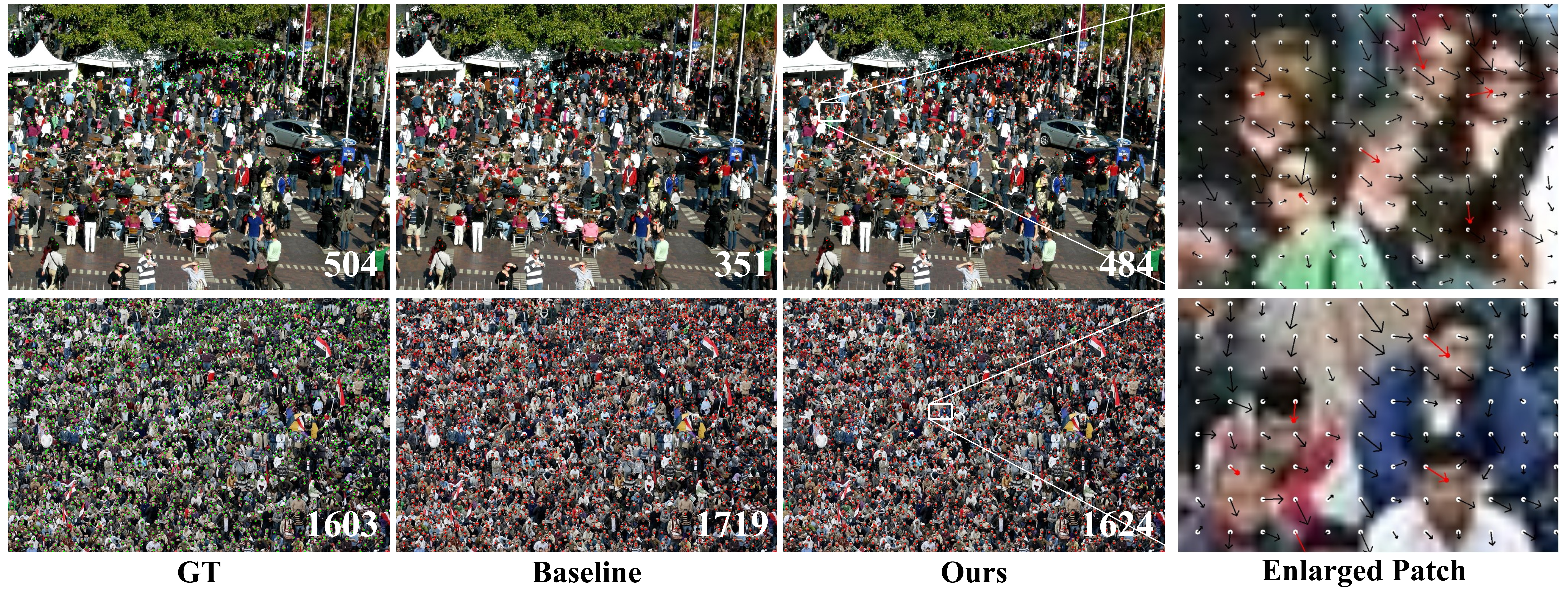}
  \caption{Some qualitative results of Baseline and our \textit{Consistent-Point}. Green points represent ground truth points, while red dots indicate the predicted points. The white numbers in the bottom-right corner of the images denote the total count of corresponding points.
}
\label{fig:qualitative_results}
    
\end{figure*}

\noindent
\textbf{Localization Metrics.}
Following \cite{wang2020nwpu, liang2022focal_FIDTM, APGCC}, we use precision (P), recall (R), and F1-measure (F) to evaluate crowd localization performance. A predicted point is considered a True Positive (TP) if its distance from the corresponding ground truth point (GT) is within a specified threshold $\sigma$; otherwise, it is considered a False Positive (FP). An unmatched ground truth point is considered a False Negative (FN).
The performance of crowd localization is given by:
$\text{P} = \frac{\text{TP}}{\text{TP} + \text{FP}}, \quad \text{R} = \frac{\text{TP}}{\text{TP} + \text{FN}}, \quad \text{F} = 2 \cdot \frac{\text{P} \cdot \text{R}}{\text{P} + \text{R}}$.
We apply fixed thresholds of $\sigma_l = 8$ and $\sigma_s = 4$ to all datasets.

\noindent
\textbf{Counting Metrics.}
We follow the standard evaluation metrics for crowd counting evaluation: Mean Absolute Error (MAE) and Mean Squared Error (MSE), defined as $\text{MAE} = \frac{1}{N_I} \sum\limits_{i=1}^{N_I} \left| y_i - \hat{y}_i \right|, \quad \text{MSE} = \sqrt{\frac{1}{N_I} \sum\limits_{i=1}^{N_I} \left(  y_i - \hat{y}_i \right)^2}$, 
where $N_I$ is the number of images in the test set, and $\hat{y}_i$ and $y_i$ denote the predicted count and corresponding ground-truth count for the $i$-th test image. 
Lower values for both metrics indicate better performance.

\begin{table}[t]
    \centering

        \resizebox{1.\linewidth}{!}{
		\begin{tabular}{l |c | c| |c c| c c}
			\specialrule{0.2em}{0pt}{0pt}
   
\rowcolor{mygray}			&   & Labeled & \multicolumn{2}{c|}{SHA}  & \multicolumn{2}{c}{SHB}  \\
            \cline{4-7}
\rowcolor{mygray}	\multirow{-2}{*}{Method} 		& \multirow{-2}{*}{Loc.} &  Ratio &  MAE$\downarrow$ & MSE$\downarrow$ & MAE$\downarrow$ & MSE$\downarrow$ \\

			 \specialrule{0.2em}{0pt}{0pt}
GP~\cite{sindagi2020learning_GP} \scriptsize{ECCV20}   & \ding{55}   & 5\%   & 102.0 & 172.0 & 15.7   & 27.9   \\
DACount~\cite{lin2022semi_DACount}  \scriptsize{ACMMM22} & \ding{55}    & 5\%   & 85.4 & 134.5 & 12.6   & 22.8   \\
        MRL~\cite{wei2023semi_MRL}  \scriptsize{TIP23}   & \ding{55}    & 5\%   & 90.9 & 141.8 & 14.8   & 22.6   \\

        MRC-Crowd~\cite{MRC-Crowd} \scriptsize{TCSVT24}  & \ding{55}  & 5\%   & 74.8 & \underline{117.3} & 11.7   & \underline{17.8}   \\
        P$^3$Net~\cite{HuiLin_pami_P3}    \scriptsize{TPAMI25}  & \ding{55}  & 5\%   & 85.5 & 131.0 & 12.0   & 22.0   \\ 
        			 \specialrule{0.08em}{0pt}{0pt}
        OT-M~\cite{lin2023optimal_OT-M}   \scriptsize{CVPR23}   & \ding{51}     & 5\%   & 83.7 & 133.3 & 12.6   & 21.5   \\
        CU~\cite{li2023calibrating_CU} \scriptsize{ICCV23}       & \ding{51}    & 5\%   & \underline{74.48} & 127.51 & \underline{11.03}   & 20.93   \\
\rowcolor{myorange}              \textbf{Ours} & \ding{51}  & 5\%   &\textbf{70.06} &\textbf{111.38} &\textbf{9.42}  &\textbf{17.38}            \\

			 \specialrule{0.2em}{0pt}{0pt}

IRAST~\cite{liu2020semi_IRAST}\scriptsize{ECCV20} & \ding{55} & 10\%      &86.9  &148.9  &14.7   &22.9   \\
IRAST+SPN~\cite{liu2020semi_IRAST}\scriptsize{ECCV20}& \ding{55}  & 10\%  &83.9  &140.1  &-      &-      \\
GP~\cite{sindagi2020learning_GP}  \scriptsize{ECCV20}& \ding{55} & 10\%         &80.4  &138.8  &12.7   &20.4   \\
PA~\cite{xu2021crowd_PA} \scriptsize{ICCV21} & \ding{55}  & 10\%         &72.79 &111.61 &12.03  &18.70  \\
DACount~\cite{lin2022semi_DACount} \scriptsize{ACMMM22} & \ding{55}    & 10\%       &74.9  &115.5  &11.1   &19.1   \\
STC-Crowd~\cite{wang2023semi_STC-Crowd} \scriptsize{TCSVT23}  & \ding{55}  & 10\%     &72.5  &118.2  &11.7   &23.8   \\
MRL~\cite{wei2023semi_MRL}\scriptsize{TIP23} & \ding{55}    & 10\%             &80.2  &125.6  &12.1   &19.7   \\
HPS~\cite{wang2024hybrid_HPS}\scriptsize{TIP24}& \ding{55}    & 10\%             &73.9  &123.5  &10.9   &20.6   \\
MRC-Crowd~\cite{MRC-Crowd}\scriptsize{TCSVT24}    & \ding{55}   & 10\%   & \underline{67.3} & \underline{106.8} & 10.3   & 18.2   \\
P$^3$Net~\cite{HuiLin_pami_P3}   \scriptsize{TPAMI25}    & \ding{55}  & 10\%   & 72.1 & 116.4 & 10.1   & 18.2    \\

\specialrule{0.08em}{0pt}{0pt}
OT-M~\cite{lin2023optimal_OT-M}  \scriptsize{CVPR23}  & \ding{51} & 10\%           &80.1  &118.5  &10.8   &18.2   \\
CU~\cite{li2023calibrating_CU} \scriptsize{ICCV23}  & \ding{51} & 10\%             &70.76 &116.62 &9.71   &17.74  \\
SAL~\cite{scale-based_active_learning}  \scriptsize{ACMMM24} & \ding{51}& 10\%   & 69.7 & 114.5 & \underline{9.7}   & \underline{17.5}  \\
\rowcolor{myorange} \textbf{Ours} & \ding{51}  & 10\%           &\textbf{63.55}&\textbf{103.18} &\textbf{8.88}  &\textbf{16.84}            \\

			 \specialrule{0.2em}{0pt}{0pt}
            GP~\cite{sindagi2020learning_GP}   \scriptsize{ECCV20}  & \ding{55}   & 40\%   & 89.0 & - & -   & -   \\
        SUA~\cite{meng2021spatial_SUA}  \scriptsize{ICCV21}   & \ding{55}  & 40\%      & 65.8 & 121.9 & 14.1   & 20.6   \\
        DACount~\cite{lin2022semi_DACount}   \scriptsize{ACMMM22}   & \ding{55}  & 40\%   & 67.5 & 110.7 & 9.6   & 14.6   \\
        STC-Crowd~\cite{wang2023semi_STC-Crowd} \scriptsize{TCSVT23}   & \ding{55}     & 40\%   & 65.1 & 106.3 & 9.8   & 16.5   \\
        MRL~\cite{wei2023semi_MRL}   \scriptsize{TIP23}    & \ding{55}     & 40\%   & 68.3 & 111.9 & 11.0   & 17.6   \\

        MRC-Crowd~\cite{MRC-Crowd}   \scriptsize{TCSVT24}  & \ding{55}   & 40\%   & 62.1 & \textbf{95.5} & 7.8   & 13.3   \\
        P$^3$Net~\cite{HuiLin_pami_P3}   \scriptsize{TPAMI25}    & \ding{55}  & 40\%   & 63.0 & 100.9 & \underline{7.1}   & \underline{12.0}    \\

        \specialrule{0.08em}{0pt}{0pt}

                OT-M~\cite{lin2023optimal_OT-M}  \scriptsize{CVPR23}  & \ding{51}   & 40\%   & 70.7 & 114.5 & 8.1   & 13.1   \\
        CU~\cite{li2023calibrating_CU}  \scriptsize{ICCV23}    & \ding{51}    & 40\%   & 64.74 & 109.56 & 7.79   & 12.70   \\
        SAL~\cite{scale-based_active_learning}  \scriptsize{ACMMM24}    & \ding{51} & 40\%   & \underline{60.7} & \underline{97.2} & 7.9   & 12.7  \\
\rowcolor{myorange}        \textbf{Ours} & \ding{51}  & 40\%           &\textbf{57.82} &99.67 &\textbf{6.65}  &\textbf{11.20}            \\

    	\specialrule{0.2em}{0pt}{0pt}
	    \end{tabular}}
	\caption{Quantitative crowd counting comparison with state-of-the-art methods on ShanghaiTech Part A and Part B datasets~\cite{zhang2016single_SH_MCNN}.
The best are in \textbf{bold} and the second-best are \underline{underlined}.  }
\label{table: counting results of ShanghaiTech Part A and ShanghaiTech Part B}

\end{table}

\begin{table}[t]
    \centering

    \setlength{\tabcolsep}{6pt}
        \resizebox{1.\linewidth}{!}{
		\begin{tabular}{l |c| c| |c c| c c}
			\specialrule{0.2em}{0pt}{0pt}
   
\rowcolor{mygray}	  & & Labeled & \multicolumn{2}{c|}{UCF-QNRF}  & \multicolumn{2}{c}{JHU-Crowd++}  \\
            \cline{4-7}
 \rowcolor{mygray} \multirow{-2}{*}{Method}   &\multirow{-2}{*}{Loc.}	& Ratio &  MAE$\downarrow$ & MSE$\downarrow$ & MAE$\downarrow$ & MSE$\downarrow$ \\

			 \specialrule{0.2em}{0pt}{0pt}
            GP~\cite{sindagi2020learning_GP}  \scriptsize{ECCV20}   & \ding{55}     & 5\%   & 160.0 & 275.0 & -   & -   \\
        DACount~\cite{lin2022semi_DACount} \scriptsize{ACMMM22}    & \ding{55}    & 5\%   & 120.2 & 209.3 & 82.2   & 294.9   \\
        MRL~\cite{wei2023semi_MRL}   \scriptsize{TIP23}   & \ding{55}     & 5\%   & 153.6 & 264.1 & 91.8   & 325.4   \\

        MRC-Crowd~\cite{MRC-Crowd} \scriptsize{TCSVT24}  & \ding{55}   & 5\%   & \textbf{101.4} & \textbf{171.3} & \underline{76.5}   & \underline{282.7}   \\
        P$^3$Net~\cite{HuiLin_pami_P3}  \scriptsize{TPAMI25}    & \ding{55}  & 5\%   & 115.3 & 195.2 & 80.8   & 306.1   \\ 

        			 \specialrule{0.08em}{0pt}{0pt}
        OT-M~\cite{lin2023optimal_OT-M}  \scriptsize{CVPR23}   & \ding{51}    & 5\%   & 118.4 & 195.4 & 82.7   & 304.5   \\
        CU~\cite{li2023calibrating_CU}  \scriptsize{ICCV23}  & \ding{51}   & 5\%   & 117.29 & 196.06 & 82.9   & 314.0   \\
\rowcolor{myorange}      \textbf{Ours}      & \ding{51}      & 5\%   & \underline{109.44} & \underline{182.89} & \textbf{75.99}   & \textbf{280.35}   \\

			 \specialrule{0.2em}{0pt}{0pt}
           IRAST~\cite{liu2020semi_IRAST} \scriptsize{ECCV20}   & \ding{55}  & 10\%   & 135.6 & 233.4 & 86.7   & 303.4   \\
     IRAST+SPN~\cite{liu2020semi_IRAST}   \scriptsize{ECCV20}   & \ding{55}   & 10\%   & 128.4 & 225.3 & -   & -   \\
            PA~\cite{xu2021crowd_PA}  \scriptsize{ICCV21}     & \ding{55}   & 10\%   & 128.13 & 218.05 & 129.65   & 400.47   \\
        DACount~\cite{lin2022semi_DACount} \scriptsize{ACMMM22} & \ding{55}    & 10\%   & 109.0 & 187.2 & 75.9   & 282.3   \\
        STC-Crowd~\cite{wang2023semi_STC-Crowd} \scriptsize{TCSVT23}   & \ding{55} & 10\%   & 123.7 & 200.9 & 103.1   & 324.7   \\
        MRL~\cite{wei2023semi_MRL}  \scriptsize{TIP23}   & \ding{55}     & 10\%   & 132.5 & 221.2 & 80.1   & 299.9   \\

        HPS~\cite{wang2024hybrid_HPS}  \scriptsize{TIP24}   & \ding{55}    & 10\%   & 121.4 & 209.6 & 100.5   & 363.4   \\
        MRC-Crowd~\cite{MRC-Crowd}\scriptsize{TCSVT24}  & \ding{55}  & 10\%   & \underline{93.4} & \underline{153.2} & 70.7   & \underline{261.3}   \\
        P$^3$Net~\cite{HuiLin_pami_P3} \scriptsize{TPAMI25}    & \ding{55}  & 10\%   & 103.4 & 179.0 & 71.8   & 294.4   \\ 

        			 \specialrule{0.08em}{0pt}{0pt}

                OT-M~\cite{lin2023optimal_OT-M} \scriptsize{CVPR23}  & \ding{51}  & 10\%   & 113.1 & 186.7 &73.0   & 280.6   \\
        CU~\cite{li2023calibrating_CU} \scriptsize{ICCV23}  & \ding{51}   & 10\%   & 104.04 & 164.25 & 74.87   & 281.69   \\
        SAL~\cite{scale-based_active_learning} \scriptsize{ACMMM24} & \ding{51} & 10\%   & 106.7 & 171.3 & \underline{69.7}   & 263.5  \\

\rowcolor{myorange}       \textbf{Ours}         & \ding{51}    & 10\%   & \textbf{89.64} & \textbf{150.63} & \textbf{69.44}   & \textbf{258.32}   \\

\specialrule{0.2em}{0pt}{0pt}
           IRAST~\cite{liu2020semi_IRAST}  \scriptsize{ECCV20}  & \ding{55}    & 40\%   & 138.9 & - & -   & -   \\
            GP~\cite{sindagi2020learning_GP}  \scriptsize{ECCV20} & \ding{55} & 40\%     & 136.0 & - & -   & -   \\
        SUA~\cite{meng2021spatial_SUA}  \scriptsize{ICCV21}    & \ding{55}   & 40\%      & 130.3 & 226.3 & 80.7   & 290.8   \\
        DACount~\cite{lin2022semi_DACount} \scriptsize{ACMMM22}  & \ding{55}   & 40\%   & 91.1 & 153.4 & 65.1   & 260.0   \\
        STC-Crowd~\cite{wang2023semi_STC-Crowd} \scriptsize{TCSVT23} & \ding{55}   & 40\%   & 98.7 & 175.6 & 72.8   & 280.4   \\
        MRL~\cite{wei2023semi_MRL}  \scriptsize{TIP23}    & \ding{55}    & 40\%   & 126.7 & 209.7 & 68.4   & 279.6   \\

        MRC-Crowd~\cite{MRC-Crowd}  \scriptsize{TCSVT24}  & \ding{55}  & 40\%   & \underline{81.1} & \textbf{131.5} & 60.0   & \textbf{227.3}   \\
        P$^3$Net~\cite{HuiLin_pami_P3} \scriptsize{TPAMI25}   & \ding{55}  & 40\%   & 90.0 & 155.4 & \textbf{58.9}   & 251.9   \\ 

        			 \specialrule{0.08em}{0pt}{0pt}

        OT-M~\cite{lin2023optimal_OT-M} \scriptsize{CVPR23}   & \ding{51}   & 40\%   & 100.6 & 167.6 & 72.1   & 272.0   \\
        CU~\cite{li2023calibrating_CU} \scriptsize{ICCV23}  & \ding{51}   & 40\%   & 91.73 & 150.79 & 66.59   & 277.43   \\
        SAL~\cite{scale-based_active_learning}  \scriptsize{ACMMM24}   & \ding{51} & 40\%   & 95.0 & 167.0 & 62.0   & 255.1  \\
\rowcolor{myorange}       \textbf{Ours}     & \ding{51}   & 40\%   & \textbf{78.91} & \underline{134.86} & \underline{59.2}   & \underline{232.4}   \\

    	\specialrule{0.2em}{0pt}{0pt}
	    \end{tabular}}
	\caption{Quantitative crowd counting comparison with state-of-the-art methods on the UCF-QNRF~\cite{idrees2018composition_UCF-QNRF} and JHU-Crowd++~\cite{sindagi2020jhu} datasets. The best are in \textbf{bold} and the second-best are \underline{underlined}. 
 }

\label{table: counting results of UCF-QNRF and JHU-Crowd++}

\end{table}

\noindent
\textbf{Implementation details.}
For a fair comparison, we follow the experimental setting in CU~\cite{li2023calibrating_CU} and SAL~\cite{scale-based_active_learning}.
Detailed implementation details including time efficiency analysis are provided in the supplementary materials.
All the results from the original paper are directly taken from it. If not available, they are reproduced using the official code based on the original paper.

\begin{table}[ht]
    \centering

        \resizebox{1.\linewidth}{!}{
		\begin{tabular}{l |c || c c |c c| c c}
			\specialrule{0.2em}{0pt}{0pt}
   
 \rowcolor{mygray} &  & \multicolumn{2}{c|}{5\%}  & \multicolumn{2}{c|}{10\%}  & \multicolumn{2}{c}{40\%} \\
            \cline{3-8}
 \rowcolor{mygray}			\multirow{-2}{*}{Method}			& \multirow{-2}{*}{Loc.}&  MAE$\downarrow$ & MSE$\downarrow$ & MAE$\downarrow$ & MSE$\downarrow$ & MAE$\downarrow$ & MSE$\downarrow$  \\

			 \specialrule{0.2em}{0pt}{0pt}

SUA~\cite{meng2021spatial_SUA}  \scriptsize{ICCV21} & \ding{55}      & - & - & -   & - & 111.7& 443.2  \\
MRL~\cite{wei2023semi_MRL} \scriptsize{TIP23}& \ding{55}    & 148.6 & 622.1 & 132.9 & 511.3 & 97.0 & 413.5 \\
HPS~\cite{wang2024hybrid_HPS} \scriptsize{TIP24} & \ding{55}          &- &-   &115.3  &532.3  &82.7   &521.1   \\
MRC-Crowd~\cite{MRC-Crowd} \scriptsize{TCSVT24}  & \ding{55}   & \textbf{95.3} & \underline{567.8} & \underline{88.1}  & 489.0  & \underline{68.2} & \underline{301.4}  \\
P$^3$Net~\cite{HuiLin_pami_P3}  \scriptsize{TPAMI25}   & \ding{55}   & 116.7 & 598.8 & 88.2   & 515.9  & 76.3 & 422.8\\

\specialrule{0.08em}{0pt}{0pt}
OT-M~\cite{lin2023optimal_OT-M} \scriptsize{CVPR23} & \ding{51}   & 112.2 & 606.3 & 104.5 & 590.5& 96.4 & 417.2       \\
CU~\cite{li2023calibrating_CU} \scriptsize{ICCV23}  & \ding{51}     &113.68 & 577.0 & 108.78 & 458.02 &88.98 &408.70 \\
SAL~\cite{scale-based_active_learning}\scriptsize{ACMMM24} & \ding{51} & - & -& 107.1 & \underline{443.6} & 92.8 & 406.7 \\
\rowcolor{myorange} \textbf{Ours} &\ding{51} &\underline{97.56} &\textbf{434.73}&\textbf{85.14}&\textbf{425.50}&\textbf{58.75} &\textbf{240.76}          \\

    	\specialrule{0.2em}{0pt}{0pt}
	    \end{tabular}}
	\caption{Quantitative crowd counting comparison with state-of-the-art methods on the NWPU-Crowd dataset~\cite{wang2020nwpu}.
The best are in \textbf{bold} and the second-best are \underline{underlined}.  }
\label{table: counting results of NWPU}

\end{table}

\begin{table*}[t]
\centering
\begin{minipage}{0.36\textwidth}
\centering
\resizebox{.9\linewidth}{!}{
\begin{tabular}{ l | c| |c c}

\specialrule{0.12em}{0pt}{0pt}            
\rowcolor{mygray}         & Labeled & \multicolumn{2}{c}{SHA}  \\
\cline{3-4}
\rowcolor{mygray} \multirow{-2}{*}{Method}	&  Ratio &MAE$\downarrow$ & MSE$\downarrow$ \\
\specialrule{0.12em}{0pt}{0pt}
Labeled only & 10\% & 78.12 & 127.12 \\
\specialrule{0.12em}{0pt}{0pt}
Baseline    & 10\%    &81.58	&139.66    \\
+ PA      & 10\%           &66.06  & 112.29  \\
+ PA + IUC      & 10\%           &\textbf{63.55}	&\textbf{103.18}  \\
\specialrule{0.12em}{0pt}{0pt}
\end{tabular}}
\caption{Ablation study on our position aggregation (PA) and instance-wise uncertainty calibration (IUC).}
\label{tab:ablation_component}
\end{minipage}%
\hspace{0.015\textwidth}  
\begin{minipage}{0.30\textwidth}
\centering
\resizebox{\linewidth}{!}{%
\begin{tabular}{c | c|| c c}
\specialrule{0.12em}{0pt}{0pt}
\rowcolor{mygray} & Labeled & \multicolumn{2}{c}{SHA}  \\
\cline{3-4}
\rowcolor{mygray}  \multirow{-2}{*}{$K$}&  Ratio & MAE$\downarrow$ & MSE$\downarrow$ \\
\specialrule{0.12em}{0pt}{0pt}
$0=0^2$  & 10\% & 67.88 & 113.51 \\
$4=2^2$  & 10\% & \textbf{63.55} & \textbf{103.18} \\
$16=4^2$ & 10\% & 70.86 & 115.70 \\
\specialrule{0.12em}{0pt}{0pt}
\end{tabular}
}
\caption{Ablation study on the number of auxiliary points in position aggregation (PA) module.}
\label{tab: ablation study number of auxiliary points}
\end{minipage}%
\hspace{0.015\textwidth} 
\begin{minipage}{0.30\textwidth}
\centering
\resizebox{1.\linewidth}{!}{
\begin{tabular}{ l | c| |c c}
\specialrule{0.12em}{0pt}{0pt}            
\rowcolor{mygray}         & Labeled & \multicolumn{2}{c}{SHA}  \\
\cline{3-4}
\rowcolor{mygray} \multirow{-2}{*}{Method}	&  Ratio &MAE$\downarrow$ & MSE$\downarrow$ \\
\specialrule{0.12em}{0pt}{0pt}
Baseline    & 10\%    &81.58	&139.66    \\
\specialrule{0.12em}{0pt}{0pt}
 + MB   & 10\%              &  79.32 &  136.23                   \\
 + TE      & 10\%           &  71.45      &        120.97        \\
 + PA      & 10\%           &\textbf{66.06}	&\textbf{112.29}  \\
\specialrule{0.12em}{0pt}{0pt}
\end{tabular}}
\caption{Ablation study on alternative solutions for PA. MB: multiple regression branches; TE: temporal ensembling.}
\label{tab: alternative solutions for PA}
\end{minipage}%
\end{table*}

\subsection{Evaluation on Crowd Localization}
\label{subsec: Evaluation on Crowd Localization}
We evaluate our method \textit{Consistent-Point} on five commonly used datasets and compare it with the state-of-the-art semi-supervised localization-based methods. 
As shown in Tab.~\ref{table: localization results under large threshold},
our \textit{Consistent-Point} consistently achieves the best localization performance across different labeled ratios ($5\%$, $10\%$, $40\%$) and different threshold values (4 or 8) of the localization metric on all five datasets. This demonstrates the effectiveness and generalizability of \textit{Consistent-Point} in various semi-supervised crowd localization scenarios and settings. 
Some qualitative results are illustrated in Fig.~\ref{fig:qualitative_results}.

\begin{figure}[th]
\centering
\includegraphics[width=.82\linewidth]{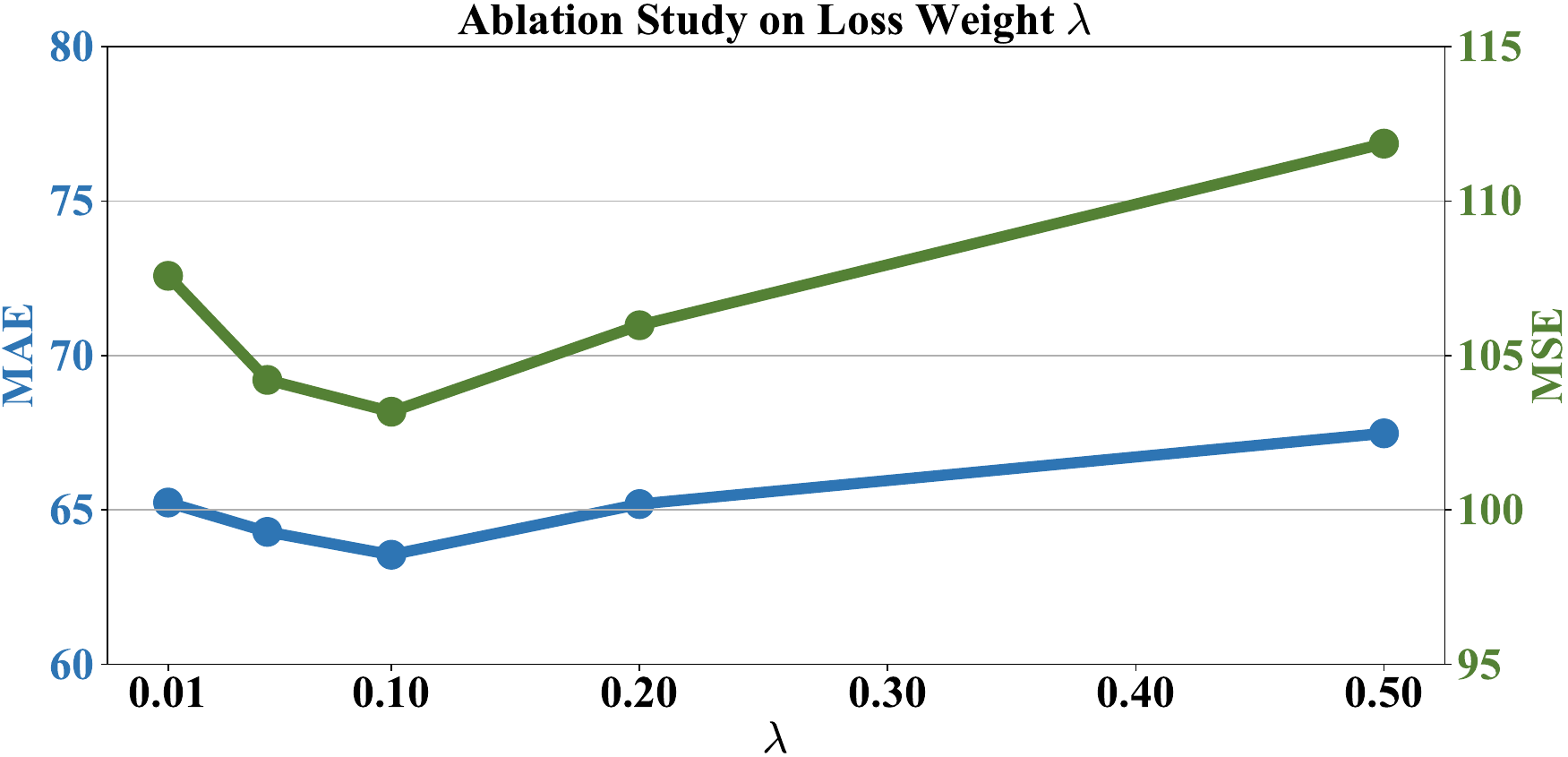}
\caption{Ablation study on the unlabeled loss weight $\lambda$.
}
\label{fig: unlabeled weight}
\end{figure}

\subsection{Evaluation on Crowd Counting}
\label{subsec: Evaluation on Crowd Counting}
To provide a comprehensive evaluation of our counting performance, we compare our method on these five datasets not only with localization-based SOTA methods but also with density-based SOTA methods. As shown in Tab.~\ref{table: counting results of ShanghaiTech Part A and ShanghaiTech Part B}, \ref{table: counting results of UCF-QNRF and JHU-Crowd++} and~\ref{table: counting results of NWPU}, we achieve new state-of-the-art results in terms of MAE across all labeled ratio settings among localization-based methods. 
When compared to density-based methods, our approach also presents impressive counting results. 
Notably, on the UCF-QNRF dataset, our method achieves better counting performance under the 40\% labeled data setting (MAE: 78.91, MSE: 134.86) than our localization-based method, P2PNet~\cite{song2021rethinking_P2P}, under the fully supervised setting with 100\% annotations (MAE: 85.32, MSE: 154.50). This improvement may be attributed to annotation errors present in the dataset.

\subsection{Ablation Study}
\label{subsec: abation study}
Ablation studies are conducted on the ShanghaiTech Part A dataset~\cite{zhang2016single_SH_MCNN} using 10\% labeled ratio setting. 
All the hyperparameters within the teacher and student networks (P2PNet) are kept fixed to ensure a fair comparison. Indeed, we only introduce two new hyperparameters: the loss weight $\lambda$ of the unlabeled data, and the number of auxiliary points, $K$, in the position aggregation module.

\noindent
\textbf{Different components.}
We first conduct an ablation study on the position aggregation (PA) and instance-wise uncertainty calibration (IUC) of our \textit{Consistent-Point}. As shown in Tab.~\ref{tab:ablation_component}, our proposed PA significantly improves the performance of the baseline method Mean-Teacher~\cite{tarvainen2017mean_teacher}. Precisely, the proposed PA achieves 15.52 MAE and 27.37 MSE improvement. Adding IUC module leads to further improvement, validating the effectiveness of different components in our \textit{onsistent-Point}. 
Our modules are also effective for Cross Pseudo Supervision (CPS)~\cite{chen2021semi_CPS} paradigm.
Detailed analysis and quantitative results are shown in the supplementary materials.

\noindent
\textbf{Loss weight $\lambda$.} We then conduct ablation study on the unlabeled loss coefficient $\lambda$ involved in Eq.~\eqref{eq:loss}. As depicted in Fig.~\ref{fig: unlabeled weight}, using different values for $\lambda$ all achieves significant improvements over the baseline counterpart (Tab.~\ref{tab:ablation_component}). The optimal performance is achieved when $\lambda$ is set to $0.1$.

\noindent
\textbf{Number of auxiliary points in position aggregation.} 
We conduct an ablation study on the number of auxiliary points in the proposed PA module. The experimental results in Tab.~\ref{tab: ablation study number of auxiliary points} show that the best performance is achieved when the number of auxiliary points is set to 4. However, when the number of auxiliary points is increased to 16 by expanding their selection range, the model performance decreases. This decline may be due to the inclusion of auxiliary points with low feature similarity. Aggregating the positional information of these points could reduce the position consistency of the pseudo-points, thereby negatively affecting the training process.

\noindent
\textbf{Comparison of alternative solutions for PA. }
PA mitigates the position fluctuation issue of pseudo-points by buffering with the positional information from surrounding auxiliary points. There are two alternatives: 1) temporal ensembling (TE) and 2) multiple regression branches (MB). 
TE aggregates pseudo-points from different time steps during training. This way requires substantial storage to save predictions from multiple epochs for all the unlabeled data and performs worse than ours (refer to Tab.~\ref{tab: alternative solutions for PA}). 
This may be because there is a certain gap between the prediction quality of the earlier epoch model and the current model. Merging their results might instead have a negative impact.
The second alternative is adding multiple regression branches (MB) and averaging their localization predictions at the same position. This approach yields results close to the baseline (Tab.~\ref{tab: alternative solutions for PA}), likely due to the identical supervision signals across these branches causing the modules to converge and produce nearly identical auxiliary points.

\section{Conclusion}
\label{sec:conclusion}
In this paper, we propose a novel point-localization-based semi-supervised crowd counting method termed \textit{Consistent-Point}, which aims to optimize the consistency of pseudo-points. 
By aggregating the positions of auxiliary points, we enhance the position consistency of pseudo-points.
Additionally, we introduce an instance-wise uncertainty calibration to improve the class consistency of pseudo-points, resulting in consistent pseudo-points for unlabeled images.
These pseudo-points provide stable supervision of high quality for the model training. 
Extensive experiments on five common datasets demonstrate the effectiveness of our method. 
Extensive experiments across five common datasets and three different labeled ratio settings demonstrate that our method achieves state-of-the-art crowd localization while also attaining impressive crowd counting results.
We hope this work will highlight the importance of pseudo-point consistency in semi-supervised crowd counting within the community.

\section{Supplementary}

\subsection{Implementation details.}
For a fair comparison, we follow the experimental setting in CU~\cite{li2023calibrating_CU} and SAL~\cite{scale-based_active_learning}.
Specifically, 
we first perform random scaling on input images within the range $[0.7, 1.3]$ and randomly crop patches of $128 \times 128$ pixels from these scaled images. 
The cropped patches are subject to random horizontally flipping with a probability of $0.5$ to further augment the dataset. 
For the datasets with very high resolution images, \textit{i.e.}, UCF-QNRF~\cite{idrees2018composition_UCF-QNRF}, JHU-Crowd++~\cite{sindagi2020jhu}, and NWPU-Crowd~\cite{wang2020nwpu}, following P2PNet~\cite{song2021rethinking_P2P} and CU~\cite{li2023calibrating_CU}, we rescale them so the longer sides of the images do not exceed 1408, 1430, and 1920, respectively.
The model is optimized using the Adam optimizer with learning rate set to $10^{-5}$ for the backbone and $10^{-4}$ for the remaining model's parameters. The coefficient for the loss weight $\lambda$ in the Eq.~10 is set to $0.1$. The training proceeds with a batch size of $8$ for each dataset on an RTX 3090 GPU using the PyTorch framework. 
We analyze the time efficiency of our method on the JHU-Crowd++ dataset~\cite{sindagi2020jhu} under 10\% labeled ratio, which contains 2272 training images and 1600 test images, with an average resolution of $1430 \times 910$. 
The training process (for 1000 epochs, same as CU~\cite{li2023calibrating_CU}) takes approximately 10 hours, while the testing process requires only 28 seconds (57.1 FPS). Notably, our method does not introduce any additional computational cost during inference, and the increase in training time is negligible (less than 2\%).

\subsection{Effectiveness of our method on CPS framework.}
To demonstrate the versatility of the proposed method, in addition to the Mean-Teacher~\cite{tarvainen2017mean_teacher} framework, we further validate the effectiveness of our \textit{Consistent-Point} with another common semi-supervised framework, Cross Pseudo Supervision (CPS)~\cite{chen2021semi_CPS}. As shown in Tab.~\ref{tab: ablation study other semi-supervised paradigm}, our PA module significantly improves the performance of the baseline using CPS. Specifically, the proposed PA achieves 15.01 MAE and 27.83 MSE improvement. The additional integration of the proposed IUC module leads to further improvements, validating the effectiveness of different proposed components. Notably, the performance of the baseline using the CPS framework, similar to that using Mean-Teacher, is poorer than the baseline trained solely on labeled data. This is primarily due to the position inconsistency and the class inconsistency of the predicted pseudo-points, which confuses the model training. This further demonstrates the importance and effectiveness of our method. The effectiveness across different pseudo-labeling semi-supervised frameworks (Mean-Teacher and CPS) confirms the versatility of our approach.

\begin{table}[h]
\centering
\setlength{\tabcolsep}{6pt}
\resizebox{0.7\linewidth}{!}{%
\begin{tabular}{l | c|| c c}
\specialrule{0.1em}{0pt}{0pt}
\rowcolor{mygray} & Labeled & \multicolumn{2}{c}{SHA}  \\
\cline{3-4}
\rowcolor{mygray} \multirow{-2}{*}{Method} & Ratio & MAE$\downarrow$ & MSE$\downarrow$ \\
\specialrule{0.1em}{0pt}{0pt}
Labeled only & 10\% & 78.12 & 127.12 \\
\specialrule{0.1em}{0pt}{0pt}
CPS (P2PNet) & 10\% & 85.86 & 146.11 \\
+ PA & 10\% & 70.85 & 118.28 \\
+ PA + IUC & 10\% & \textbf{65.74} & \textbf{106.27} \\
\specialrule{0.1em}{0pt}{0pt}
\end{tabular}
}
\caption{Ablation study on the proposed position aggregation (PA) and instance-wise uncertainty calibration (IUC) based on the CPS~\cite{chen2021semi_CPS} framework with ShanghaiTech Part A dataset~\cite{zhang2016single_SH_MCNN}.}
\label{tab: ablation study other semi-supervised paradigm}  
\end{table}

\subsection{Limitation}
\label{subsec: limitation}
In a few extremely dense regions, the PA module may take pseudo-points belonging to other heads as auxiliary neighboring points, which can somewhat increase position inconsistency. In future work, we will consider dynamically adjusting the range and number of auxiliary points based on density information.

{
    \small
    \bibliographystyle{ieeenat_fullname}
    \bibliography{main}
}

\end{document}